\documentclass[letterpaper, 10 pt, conference]{ieeeconf}  

\IEEEoverridecommandlockouts                              

\usepackage{algorithm}
\usepackage{algorithmic}
\usepackage[algo2e]{algorithm2e}
\usepackage{cite}
\usepackage{amsmath}
\usepackage{graphicx} 
\usepackage{lipsum}    
\usepackage{float}     
\usepackage{caption}   
\usepackage{booktabs}
\usepackage{amssymb}
\title{\Large \bf
    FODMP: \underline{F}ast \underline{O}ne-Step \underline{D}iffusion of \underline{M}ovement \underline{P}rimitives Generation for Time-Dependent Robot Actions
}

\author{Xirui Shi$^{1,\dag}$, Arya Ebrahimi$^{1}$, Yi Hu$^{1}$, and Jun Jin$^{1,2}$
\thanks{$^{1}$The authors are with the Department of Electrical and Computer Engineering, University of Alberta, Edmonton, Alberta, Canada.}
\thanks{$^{2}$Jun Jin is also with the Alberta Machine Intelligence Institute(Amii), Edmonton, Alberta, Canada.}%
\thanks{\dag Corresponding to: Xirui Shi (dalen.shi@ualberta.ca)}
}

\begin{document}

\maketitle
\thispagestyle{empty}
\pagestyle{empty}

\begin{abstract}
Diffusion models are increasingly used for robot learning, but current designs face a clear trade-off. Action-chunking diffusion policies like ManiCM are fast to run, yet they only predict short segments of motion. This makes them reactive, but unable to capture time-dependent motion primitives --- such as following a spring–damper-like behavior with built-in dynamic profiles of acceleration and deceleration. Recently, Movement Primitive Diffusion (MPD) partially addresses this limitation by parameterizing full trajectories using Probabilistic Dynamic Movement Primitives (ProDMPs), thereby enabling the generation of temporally structured motions. Nevertheless, MPD integrates the motion decoder directly into a multi-step diffusion process, resulting in prohibitively high inference latency that limits its applicability in real-time control settings.
We propose FODMP (Fast One-step Diffusion of Movement Primitives), a new framework that distills diffusion models into the ProDMPs trajectory parameter space and generates motion using a single-step decoder. FODMP retains the temporal structure of movement primitives while eliminating the inference bottleneck through single-step consistency distillation. This enables robots to execute time-dependent primitives at high inference speed, suitable for closed-loop vision-based control.
On standard manipulation benchmarks (MetaWorld, ManiSkill), FODMP runs up to 10× faster than MPD and 7× faster than action-chunking diffusion policies, while matching or exceeding their success rates.
Beyond speed, by generating fast acceleration-deceleration motion primitives, FODMP allows the robot to intercept and securely catch a fast-flying ball, whereas action-chunking diffusion policy and MPD respond too slowly for real-time interception.

\end{abstract}
\section{Introduction}
Diffusion models have recently emerged as one of the most popular choices for generating high-dimensional, continuous robot actions, particularly in vision--language--action (VLA) frameworks~\cite{he2023diffusion,black2024pi_0, ze20243d, shi2025hi}. Unlike autoregressive token-based approaches, diffusion models can directly capture multi-modal action distributions in high degrees of freedom (DoF), making them attractive for large-scale robot learning \cite{chi2023diffusion,chi2023diffusionpolicy,pearce2023imitating,reuss2024multimodal}. This trend is already visible in recent VLA systems such as $\pi_0$ \cite{black2024pi_0}, which integrate diffusion-based action generators into generalist robot policies. 

\begin{figure}
    \centering
    \includegraphics[width=1\linewidth]{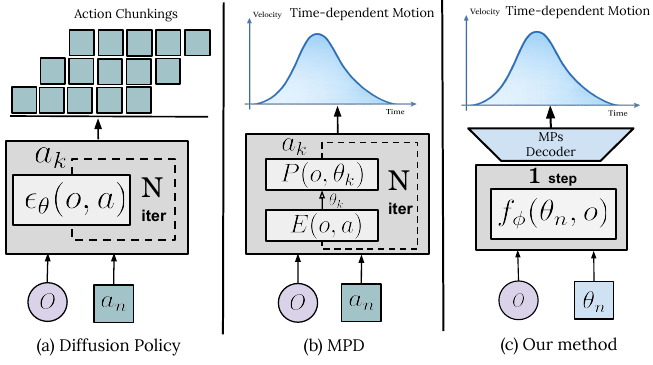}
    \caption{\textbf{Comparison of Policy Designs. }(a) Action chunking–based diffusion policies predict discrete sequences of actions without explicitly modeling time-dependent motion, resulting in temporally unstructured behavior. (b) MPD introduces a motion decoder into the diffusion process to recover time-dependent trajectories; however, motion parameters are still inferred through multi-step denoising, leading to high inference latency. (c) Our method directly predicts movement primitive parameters via one-step diffusion, followed by a single decoder pass to generate time-dependent motion. This design enables low-latency inference while explicitly modeling continuous motion evolution over time.}
    \label{fig:1}
\end{figure}

However, for diffusion to become a \emph{general motion generation backbone} in embodied AI, it must go beyond simple reaching or pick-and-place behaviors~\cite{VLA_RT2} and be capable of producing a wide range of temporally structured robot motions required in real-world tasks \cite{dasari2024ingredients,carvalho2024motion}. Such motions include anticipatory interception behaviors that tightly couple acceleration and deceleration to match the trajectory of a flying ball, enabling a smooth transition from rapid approach to controlled momentum absorption at contact. The ability to generate these structured, time-dependent interception primitives significantly expands the capability of the action generation module: rather than reacting with short, myopic control steps, the policy can plan and execute coherent interception trajectories over time, enabling reliable ball catching under fast dynamics and limited reaction windows.

Two dominant paradigms exist for applying diffusion models to robot action generation. The first is \emph{action chunking}, where the policy predicts short horizons of low-level actions or waypoints through diffusion \cite{chi2023diffusion,chi2023diffusionpolicy,pearce2023imitating,reuss2024multimodal,he2023diffusion}. These approaches are efficient and widely adopted in VLA pipelines, but their reliance on short horizons makes them inherently reactive and unable to produce temporally coherent behaviors. In contrast, \emph{Movement Primitive Diffusion} (MPD) \cite{scheikl2024movement} generates trajectories by diffusing over Probabilistic Dynamic Movement Primitives (ProDMPs) \cite{li2023prodmp,saveriano2023dynamic,ijspeert2013dynamical,schaal2005learning,paraschos2013probabilistic}, capturing temporal structure and complex profiles. However, MPD sacrifices the efficiency advantage of action chunking: its multi-step denoising is too slow for closed-loop deployment. In short, action chunking trades away expressiveness for speed, while MPD trades away speed for expressiveness, leaving a gap for methods that can achieve both.

Bridging this gap is critical if diffusion is to serve as a \emph{general motion generation backbone} in embodied AI. Real-world skills often depend on \emph{time-dependent motion primitives}. These primitives are fundamental building blocks of robot behavior, yet they cannot be adequately represented by action-chunking diffusion policies, which reduce motions to short reactive segments, nor can they be deployed effectively with MPD due to its latency. 

Classic methods like Dynamic Movement Primitives (DMPs) and Probabilistic Movement Primitives (ProMPs) have long emphasized the role of temporal structure in motor generation \cite{ijspeert2013dynamical,schaal2005learning,paraschos2013probabilistic,li2023prodmp,saveriano2023dynamic}, and recent diffusion-based extensions~\cite{Scheikl2024MPD} confirm their potential for representing rich robot motions. However, an open challenge remains: how to combine the temporal expressiveness of movement primitives with the efficiency required for closed-loop diffusion policies.

We address this challenge with \emph{FODMP} (Fast One-step Diffusion of Movement Primitives), a framework that unifies movement primitive representations with one-step diffusion via consistency distillation. Instead of relying on multi-step denoising, FODMP directly maps noise to ProDMPs trajectory parameters in a single inference step, thereby preserving temporal expressiveness while achieving closed-loop control rates. This design is inspired by recent advances in consistency models \cite{song2023consistency,song2023improved,kim2023latent,dai2024motionlcm,chen2024pixart}, which demonstrate that diffusion models can be distilled into fast one-step generators without sacrificing performance. Our contributions are as follows:
\begin{itemize}
    \item We introduce \emph{FODMP}, the first one-step diffusion framework that combines consistency distillation with ProDMPs to generate temporally expressive primitives at control-rate speeds.
    \item We demonstrate FODMP on MetaWorld \cite{yu2020meta}, ManiSkill \cite{mu2021maniskill}, and two real-world tasks: Push-T and Ball-Catching, showing that it captures motion primitives which action chunking cannot reproduce and MPD cannot deploy in time due to its slow inference.
\end{itemize}

To the best of our knowledge, \emph{FODMP} is the first one-step diffusion framework that makes structured time-dependent motions generation practical for closed-loop robot control. By combining the temporal expressiveness of ProDMPs with one-step diffusion, it overcomes the limitations of both action chunking and multi-step MPD. We envision this as a step toward more capable action generation modules in embodied AI foundation models, where diffusion serves as the action expert to generate scalable, structured motions for richer skills and more capable agents.

\begin{figure*}[htbp]
    \centering
    \includegraphics[width=\textwidth]{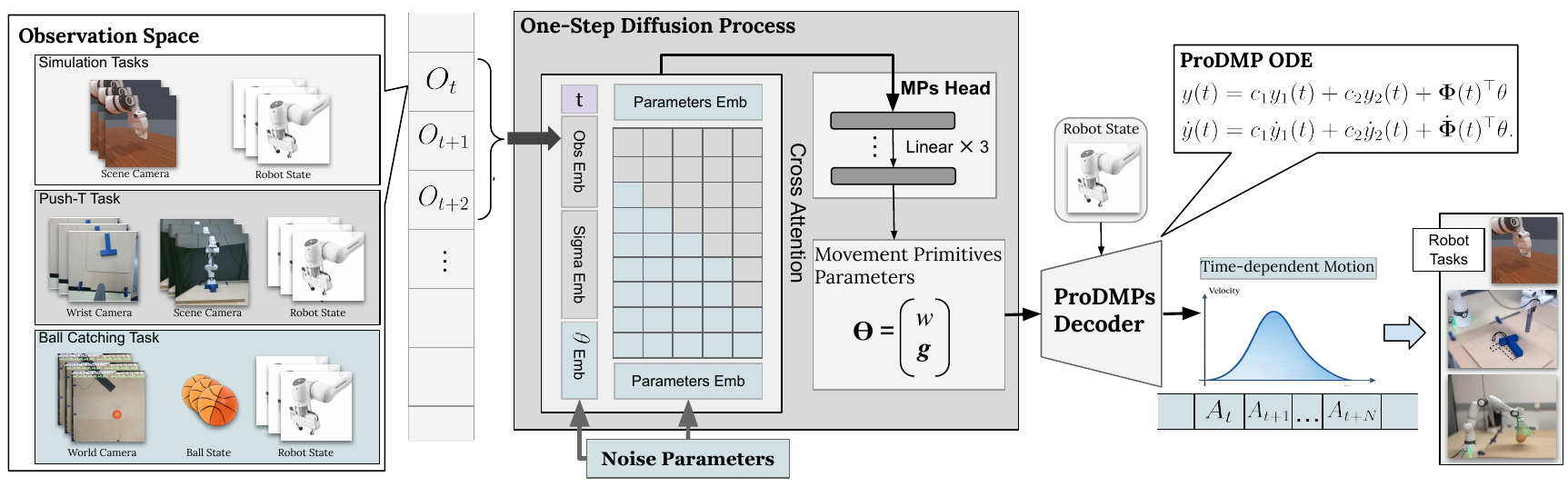}
    \caption{\textbf{Method Overview. }Our approach conditions a Transformer-based one-step diffusion model on multi-modal observations to directly predict ProDMPs parameters. The predicted parameters are decoded by a ProDMPs ODE solver, enabling fast generation of smooth, structured robot trajectories across diverse robot tasks.}
    \label{overview}
\end{figure*}


\section{Related Work}

\subsection{Diffusion Models for Robot Learning}
Diffusion models were first introduced for image generation and denoising~\cite{ho2020denoising,song2020score}, and have since been adapted for robotics as powerful action decoders. Early work showed that planning can be framed as a generative refinement process, where noisy trajectories are iteratively denoised into feasible control commands~\cite{janner2022planning,chi2023diffusion,chi2023diffusionpolicy}. Recent advances, including $\pi_0$~\cite{black2024pi_0} and multimodal diffusion transformers~\cite{reuss2024multimodal}, further established diffusion as the dominant backbone for high-DoF action generation in vision--language--action models. Despite these successes, most existing diffusion policies operate under the paradigm of \emph{action chunking}, where short horizons of actions or waypoints are predicted at each step~\cite{chi2023diffusion,pearce2023imitating,he2023diffusion}. While efficient, this design makes policies inherently reactive and unable to capture structured, time-dependent motion profiles. Our work overcomes this limitation by generating full motion primitives with built-in temporal structure, while retaining the inference efficiency of action chunking methods.

\subsection{Movement Primitives in Robotics}
Movement primitives (MPs) have long provided a compact and structured representation of robot motions. Classical Dynamic Movement Primitives (DMPs) encode trajectories through attractor dynamics with forcing functions, enabling spatial and temporal invariance but relying on numerical integration~\cite{ijspeert2013dynamical,schaal2005learning,saveriano2023dynamic}. Probabilistic Movement Primitives (ProMPs) extend this idea by representing trajectories as distributions that capture temporal correlations across degrees of freedom, making them effective for learning from demonstrations~\cite{paraschos2013probabilistic}. More recently, ProDMPs~\cite{li2023prodmp} unified dynamic and probabilistic formulations through precomputed basis functions, eliminating integration costs. Diffusion has also been applied in the trajectory-parameter space: Carvalho et al.~\cite{carvalho2024motion} applied diffusion to motion planning, and Scheikl et al.~\cite{scheikl2024movement} introduced Movement Primitive Diffusion (MPD), which diffuses over ProDMPs parameters to produce smooth and temporally consistent motions. However, MPD inherits the inefficiency of multi-step denoising, losing the speed advantage of action chunking methods. Our work differs by applying \emph{one-step consistency distillation} in the ProDMPs space, combining temporal expressiveness with control-rate efficiency.

\subsection{Accelerating Diffusion Inference}
The high cost of multi-step denoising has spurred significant research into accelerating diffusion. DPM-Solver~\cite{lu2022dpm} and related ODE solvers reduce the number of steps, while consistency models~\cite{song2023consistency,song2023improved} introduce self-consistency along the probability flow ODE trajectory, allowing a single noisy input to map directly to the final clean output. Extensions such as Latent Consistency Models (LCM)~\cite{kim2023latent,dai2024motionlcm} and PixArt-$\delta$~\cite{chen2024pixart} demonstrate large speedups in image generation. Robotic applications are emerging: ManiCM~\cite{lu2024manicm} applies consistency to visuomotor policies. However, prior work has not integrated consistency distillation with movement primitive representations. Our method closes this gap by distilling diffusion directly in the ProDMPs trajectory space, enabling fast, structured, and temporally consistent robot motion generation.

\section{Methodology}

In this section, we present our proposed approach---Fast One-step Diffusion of Movement Primitives (FODMP)---together with the necessary background.
\subsection{Problem Formulation}

Our goal is to learn a conditional generative policy that produces robot trajectories from observations, capturing both the temporal expressiveness of motion profiles and the efficiency required for real-time control. Following~\cite{scheikl2024movement}, we represent the training dataset as
\begin{equation}
    \mathcal{D} = \{ (o_j, \{\tau_{jk}\}_{k=1}^{N_j}) \}_{j=1}^N,
\end{equation}
where $o_j \in \mathcal{O}$ denotes the $j$-th observation or task specification (e.g., RGB images, proprioceptive states, start/goal poses, or environment descriptions), and $\{\tau_{jk}\}_{k=1}^{N_j}$ is the set of $N_j$ expert demonstration trajectories paired with $o_j$.

Each trajectory $\tau$ is represented by a parameter vector $\theta \in \Theta$, such that
\begin{equation}
    \tau = \text{Decode}(\theta),
\end{equation}
where the decoder depends on the choice of trajectory representation. In general, $\theta$ may correspond to (i) the weights of a neural network generating full trajectories, (ii) the coefficients of B-splines~\cite{liao2024bmp} describing geometric paths, or (iii) the parameters of movement primitives. 

In this work, we adopt the \emph{Probabilistic Dynamic Movement Primitive (ProDMPs)} formulation. Under this choice, $\Theta$ is instantiated as the ProDMPs parameter space, where $\theta = [\mathbf{w}, g]$ includes the forcing weights $\mathbf{w}$ and attractor $g$, together with constants $c_1, c_2$ determined by initial boundary conditions. This unification ensures that trajectories generated from $\theta$ encode entire time-dependent motion profiles, including acceleration and deceleration phases.

Given an observation $o \in \mathcal{O}$, the learning objective is to train a policy
\begin{equation}
    \pi: \mathcal{O} \rightarrow \Theta,
\end{equation}
that maps observations to ProDMPs parameters. We therefore aim to model the conditional distribution $p(\theta \mid o)$ and generate structured trajectories through ProDMPs decoding. The Overview of our method are showed in Fig.~\ref{overview}
\subsubsection{Probabilistic Dynamic Movement Primitives (ProDMPs)} 
\label{sec:prodmp}

Movement primitives (MPs) provide a compact and structured representation of robot trajectories. In this work, the trajectory parameter $\theta \in \Theta$ corresponds to the parameters of a Probabilistic Dynamic Movement Primitive (ProDMPs)~\cite{li2023prodmp}. Specifically,
\begin{equation}
    \theta = [\mathbf{w}, g],
\end{equation}
where $\mathbf{w}$ are the weights encoding the nonlinear forcing term that shapes the trajectory, and $g$ is the goal attractor ensuring convergence. Here $\mathbf{w}$ and $g$ are learned by networks. Constants $c_1, c_2$ are determined from the initial boundary conditions $(y_0, \dot{y}_0)$ and are not learned.

A trajectory position $y(t)$ is then expressed as
\begin{equation}
    y(t) = c_1 y_1(t) + c_2 y_2(t) + \mathbf{\Phi}(t)^\top \theta,
\end{equation}
where $y_1(t), y_2(t)$ are complementary solutions of the ProDMPs ODE, and $\mathbf{\Phi}(t)$ is the precomputed basis matrix. The corresponding velocity is
\begin{equation}
    \dot{y}(t) = c_1 \dot{y}_1(t) + c_2 \dot{y}_2(t) + \dot{\mathbf{\Phi}}(t)^\top \theta.
\end{equation}

This formulation eliminates the need for costly numerical integration required by classical DMPs, while retaining the temporal consistency and generalization ability of probabilistic MPs. Crucially, the ProDMPs parameters $\theta$ encode entire time-dependent motion profiles—such as accelerating to intercept and decelerating to stop in a spring–damper–like manner—making them particularly suitable for dynamic robotic tasks where smooth but reactive control is required.

\subsubsection{Diffusion Models}
Diffusion models generate data by inverting a forward noising process. For trajectory parameters $\theta$, the forward process is defined as:
\begin{equation}
    q(\theta_t \mid \theta_{t-1}) = \mathcal{N}(\alpha_t \theta_{t-1}, \sigma_t^2 I),
\end{equation}
which gradually perturbs clean parameters into Gaussian noise. The reverse process denoises iteratively via a learned score function. However, this multi-step denoising incurs significant latency, making real-time control infeasible.

\subsubsection{Consistency Distillation}
Consistency models~\cite{song2023consistency} accelerate sampling by enforcing a self-consistency property along the Probability Flow ODE (PF-ODE). A consistency function $f_\phi$ maps noisy inputs directly to clean outputs:
\begin{equation}
    f(\theta_t, t) = f(\theta_{t'}, t'), \quad \forall t, t' \in [\epsilon,T].
\end{equation}
This property ensures that any intermediate noisy sample can be mapped to the same clean parameter $\theta_0$, enabling single-step inference after training. Consistency distillation trains $f_\phi$ by transferring knowledge from a pre-trained multi-step teacher diffusion model.

\subsection{FODMP: Fast One-Step Diffusion of Movement Primitives}
\begin{figure}
    \centering
    \includegraphics[width=\linewidth]{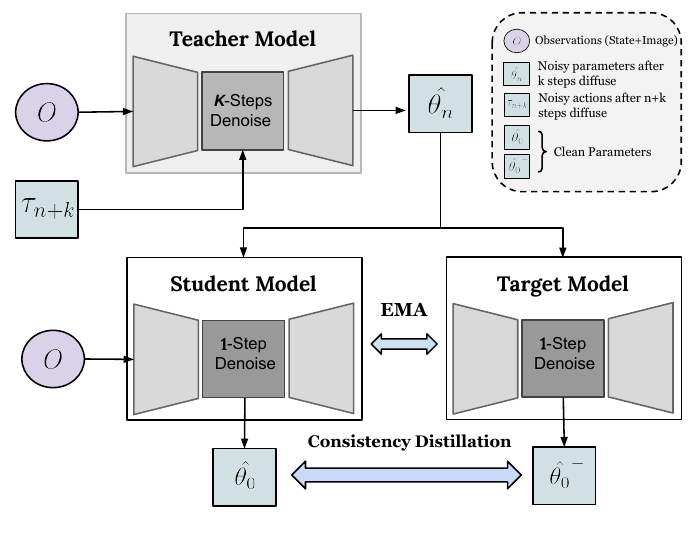}
    \caption{\textbf{Consistency Distillation Pipeline. }Given observations, a multi-step teacher performs k-step denoising to obtain cleaner parameters, which supervise a one-step student model. The student is trained using a consistency objective against a target model, whose parameters are updated via EMA from the student.}
    \label{fig:distillation pipeline}
\end{figure}
\subsubsection{Teacher Model Setup}

Following Movement Primitive Diffusion (MPD)~\cite{scheikl2024movement}, we first train a teacher diffusion model that denoises ProDMPs parameters $\theta = [\mathbf{w}, g]$, where $\mathbf{w}$ encodes the forcing term and $g$ denotes the attractor.
In MPD, the motion decoder is explicitly integrated into a multi-step diffusion process, resulting in a model that captures the distribution over action sequences.

The teacher model consists of a network $E_\vartheta$ that predicts motion parameters $\tilde{\theta}$ from noisy action samples $\tilde{\tau}$ and observations $o$:
\begin{equation}
    \tilde{\theta} = E_\vartheta(\tilde{\tau}, o, t).
\end{equation}
The predicted parameters are then decoded into a full trajectory using the ProDMPs decoder $P_\Phi$:
\begin{equation}
    F_\vartheta(\tilde{\tau}, o, t) = P_\Phi(y_0, \dot{y}_0, E_\vartheta(\tilde{\tau}, o, t)).
\end{equation}

The corresponding diffusion score function is approximated as:
\begin{equation}
    \nabla_{\theta} \log p(\tau \mid o, \sigma(t)) \approx 
    \frac{F_\vartheta(\tilde{\tau}, o, \sigma(t)) - \tau}{\sigma(t)^2}.
\end{equation}

While this teacher model generates temporally coherent trajectories, it relies on multiple denoising steps (typically 10--50), resulting in inference latency that is prohibitive for real-time deployment.
MPD repeatedly invokes the motion decoder inside the diffusion loop, whereas our method performs diffusion directly over motion parameters. Accordingly, in the distillation process, we retain only the network $E_\vartheta$ to map actions to motion parameter distributions, while performing motion decoding outside the diffusion process.

\subsubsection{Student Model via Consistency Distillation}
To eliminate the multi-step inference, we distill the teacher into a one-step student model. Specifically, we first obtain noisy parameters $\theta_{n+k}$ from the diffused noisy action $\tau_{n+k}$ at diffusion step $n+k$. Given a noisy input $\theta_{n+k}$ at diffusion step $n+k$, the teacher produces a $k$-step denoised estimate $\hat{\theta}_n$, while the student directly predicts the clean parameter $\hat{\theta}_0$. Here, $n$ denotes the current diffusion step, and $k$ is a skip interval controlling how many steps the teacher simulates. 

To enforce self-consistency, we introduce a target network $f_{\phi^-}$ updated via exponential moving average (EMA). The distillation loss is:
\begin{equation}
    \mathcal{L}_{CD} = \mathbb{E}\big[\lambda(t_n)\, d(f_\phi(\theta_{n+k}, o, t_{n+k})-f_{\phi^-}(\hat{\theta}_n, o, t_n))\big],
\end{equation}
where $d(\cdot,\cdot)$ is a distance metric and $\lambda(t)$ a weighting function. This training procedure enables the student to directly map noisy ProDMPs parameters to their clean version in a single step. Fig.~\ref{fig:distillation pipeline} and Algorithm~\ref{alg:fodmp} outlines the consistency distillation process for FODMP.

\subsubsection{Closed-Loop Control}
During inference, FODMP runs in a receding-horizon loop:
\begin{itemize}
    \item Sample $\theta_T \sim \mathcal{N}(0, I)$.
    \item Compute $\theta_0 = f_\phi(o, \theta_T, T)$.
    \item Decode the trajectory $\tau = P_\Phi(y_0, \dot{y}_0, \theta_0)$.
    \item Execute the action $\tau$, then re-plan at the next control step.
\end{itemize}
This process preserves temporal structure through ProDMPs while achieving the efficiency of one-step inference. 

\begin{algorithm}[htbp]
    \caption{Consistency Distillation for FODMP}
    \label{alg:fodmp}
    \begin{algorithmic}[1]
        \STATE Input dataset $\mathcal{D}$, parameters $\phi$, learning rate $\eta$
        \REPEAT
        \STATE Sample $o, \tau_0 \sim \mathcal{D}$ and $n \sim \mathcal{U}[1,N-k]$
        \STATE Add noise: $\tau_{n+k} \sim \mathcal{N}(\tau_0, t_{n+k}^2 I)$
        \STATE Calculate $\theta_0=E_\vartheta(\tau_0, o, t_0)$
        \STATE Calculate $\theta_{n+k} = E_\vartheta(\tau_{n+k}, o, t_{n+k})$
        \STATE Teacher $k$-step denoise: $\hat{\tau}_n \leftarrow \tau_{n+k}+(t_n - t_{n+k})\phi(\tau_{n+k}, t_{n+k})$
        \STATE Calculate $\hat{\theta_n} = E_\vartheta(\hat{\tau_n}, o, t_n)$
        \STATE Student one-step prediction: $\hat{\theta}_0 \gets f_\phi(\theta_{n+k}, o, t_{n+k})$
        \STATE Compute distillation loss $\mathcal{L}_{CD}$
        \STATE Update $\phi \gets \phi - \eta \nabla_\phi \mathcal{L}_{CD}$
        \STATE Update target network $\phi^- \gets \mu \phi^- + (1-\mu)\phi$
        \UNTIL convergence
    \end{algorithmic}
\end{algorithm}

\section{Evaluations}
\label{IV}
We evaluate the proposed FODMP on both simulation benchmarks and real-world robotic tasks to comprehensively assess its effectiveness in motion generation. Our experiments are designed to answer three key questions:

(1) whether FODMP can generate smooth and temporally consistent trajectories across diverse environments and task settings;

(2) whether FODMP enables one-step inference while preserving reliable performance compared to multi-step generative policies; and

(3) how much performance improvement FODMP achieves relative to state-of-the-art baseline methods.


\subsection{Simulation Experiments}
We conduct simulation experiments on the well-established MetaWorld~\cite{yu2020meta} and ManiSkill~\cite{Mu2021ManiSkillGM} benchmarks, which provide a diverse set of manipulation tasks with varying levels of difficulty.

\subsubsection{Tasks and Datasets} 
\begin{figure}[htbp]
    \centering
    \includegraphics[width=\linewidth]{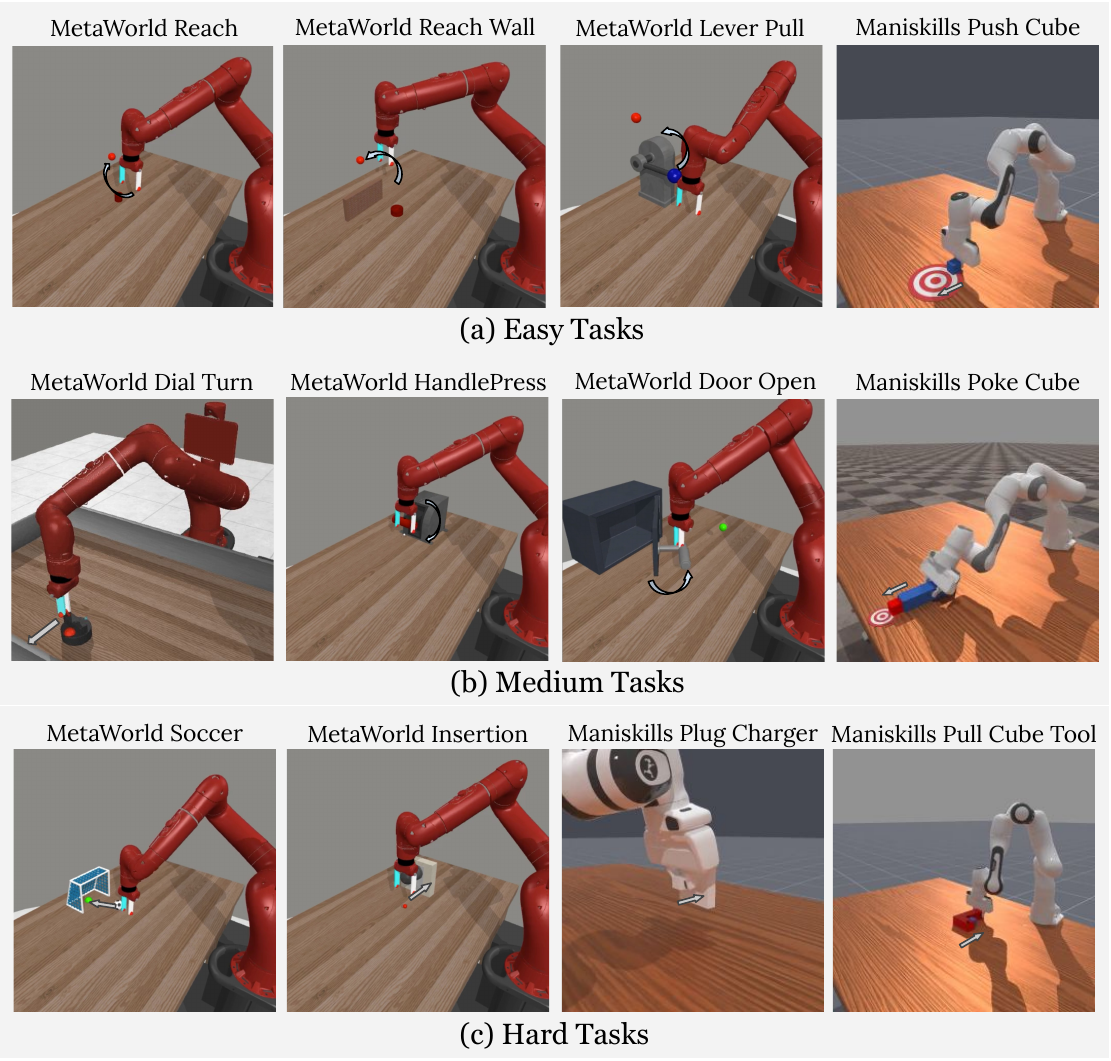}
    \caption{\textbf{Simulation Environments. }We evaluate our method on a diverse set of manipulation tasks from MetaWorld and ManiSkill, grouped by difficulty into easy, medium, and hard categories.}
    \label{fig:simulation_envs}
\end{figure}

As shown in Fig.~\ref{fig:simulation_envs}, the selected tasks are grouped into three difficulty levels: easy, medium, and hard. The easy tasks mainly involve basic reaching and pushing motions with minimal precision and reactive requirements. The medium tasks require more structured interactions, such as turning dials, pressing handles, or opening doors. The hard tasks involve precise contact or tool use, posing significant challenges for motion generation and control. For dataset collection, we follow a standard demonstration-based pipeline in each benchmark. Each demonstration consists of a sequence pair $(\tau_i, o_i)$ over a full task execution with $N$ time steps, where $\tau_i$ denotes the action trajectory and $o_i$ represents the corresponding observations.

\subsubsection{Metrics} In simulation experiments, we report two key metrics to assess the performance of our method: (1) Task Metrics, which measure the success rate on benchmark tasks. For the success rate, the higher the better. (2) Time Metrics, which evaluate the efficiency of inference. We measure the average runtime per step. For inference time, lower values indicate better performance.
\begin{table}[htbp]
    \centering
    \small
    \caption{Success rate and inference time across all tasks based on our models and baselines. The best results for each category are in bold font and the second best ones are underlined for an easier comparison.}
    \label{tab}
    \renewcommand{\arraystretch}{1.4}
    \begin{tabular}{lccccc}
        \toprule
        \textbf{Methods} & \textbf{Easy} & \textbf{Medium} & \textbf{Hard} & \textbf{Average} \\
        \midrule
        \multicolumn{5}{c}{\textbf{Success Rate (\%)}} \\
        DP  & \textbf{99.3}$\pm${0.1}  & 41.0$\pm${3.2}  & 10.1$\pm${1.4}  & 50.1 \\
        ManiCM   & 79.2$\pm${0.4} & \underline{18.9}$\pm${5.1} & \underline{5.2}$\pm${0.2} &  34.4 \\
        MPD   & 98.9$\pm${0.3} & \underline{64.8}$\pm${2.6} & \underline{28.6}$\pm${2.9} &  \underline{64.1} \\
        FODMP & \underline{99.2}$\pm${0.1} & \textbf{86.3}$\pm${1.2} & \textbf{49.0}$\pm${2.3} & \textbf{78.2}  \\
        
        \midrule
        \multicolumn{5}{c}{\textbf{Inference Time (ms)}} \\
        DP  & \underline{119.8}$\pm${1.4} & \underline{121.3}$\pm${2.3}  & \underline{118.2}$\pm${3.6} & 119.7 \\
        ManiCM  & \textbf{16.2}$\pm${0.8} & \textbf{15.6}$\pm${3.4}  & \textbf{16.9}$\pm${1.1} & \textbf{16.2} \\
        MPD  & 162.7$\pm${3.5} & 173.2$\pm${3.6} & 169.9$\pm${1.2}  & 168.6 \\
        FODMP  & \textbf{15.2}$\pm${0.8} & \textbf{18.6}$\pm${3.4}  & \textbf{17.9}$\pm${1.1} & \underline{17.2}\\
        
        \bottomrule
    \end{tabular}
\end{table}

\subsubsection{Baselines} Our work focuses on accelerating inference speed while improving action generation quality. During evaluation, we compare FODMP against Transformer-based Diffusion Policy (DP), Movement Primitive Diffusion (MPD), and ManiCM.

\subsubsection{Evaluation Methodology}
We report the best-performing results for each baseline method on each benchmark, selected from all available sources. All reported results are averaged over three independent training seeds, using the mean performance of the last 10 checkpoints (saved every 100 epochs) for each run. For evaluation, each trained policy is executed for 50 rollouts per task, and the success rate is computed by averaging across these rollouts. Each method is evaluated using its best-performing action representation: position control for DP and ManiCM, and velocity control for MPD and our methods.

\subsubsection{Results and Comparisons}

The results from these simulation benchmarks are summarized in Table.~\ref{tab}. FODMP achieves the highest average success rate (78.2\%), outperforming DP (50.1\%), ManiCM (34.4\%), and MPD (64.1\%). While all methods perform similarly on easy tasks, FODMP shows clear advantages on medium (86.3\%) and hard (49.0\%) tasks, which require structured and temporally consistent motion generation. In terms of efficiency, FODMP runs at 17.2 ms per step, enabling real-time control. Compared to MPD (168.6 ms) and DP (119.7 ms), FODMP achieves significantly lower latency without sacrificing performance, demonstrating a favourable trade-off between success rate and inference efficiency.

\begin{figure}[htbp]
    \centering
    \includegraphics[width=\linewidth]{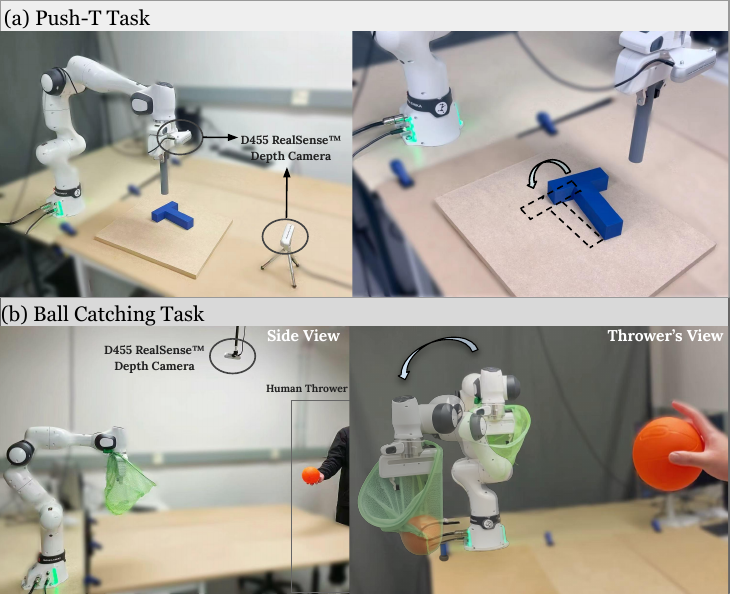}
    \caption{\textbf{RealWorld Experiments Setups.} (a) Push-T task setup, where a Franka arm pushes a T-shaped object using visual observations from two RealSense cameras. (b) Ball-catching task setup, where the robot intercepts a human-thrown ball using depth observations.}
    \label{fig:hardware}
\end{figure}

\subsection{RealWorld Push-T Experiments}

We evaluate FODMP on a real-world Push-T task to study its ability to perform precise and visually guided tabletop manipulation. The Push-T task involves interaction with a static object. As a result, the primary challenge lies in maintaining stable and time-consistent motion during continuous contact. Successful execution requires the robot to generate smooth and coherent action sequences to avoid oscillations or drift caused by temporally incoherent commands.

\subsubsection{Hardware Steup}
As showed in Fig~\ref{fig:hardware}, the Push-T task is conducted using a Franka Emika Panda arm performing planar pushing of a T-shaped object. Visual observations are provided by two Intel RealSense D455 depth cameras.

\subsubsection{Datasets}

We collect demonstration data via human teleoperation using joystick control. A total of 350 demonstrations are collected with randomized initial poses of both the T-shaped object and the robot. Each demonstration consists of paired sequences of observations and actions over a full task execution. The observation space includes robot proprioceptive states and RGB images captured from two cameras, while the action space comprises Cartesian velocity commands applied to the robot end-effector.

\subsubsection{Metrics}In addition to the evaluation metrics used in the simulation experiments, we introduce data-efficiency metrics for the Real World tasks, as collecting real world robot demonstrations via human teleoperation is particularly challenging. As a result, we assess how effectively each method leverages limited training data by evaluating performance under varying numbers of demonstrations.

\subsubsection{Evaluation Methodology}We follow the evaluation proccess introduced in \cite{chi2023diffusionpolicy}. The robot need to push the T to the target block and then move its end to end zone. In successful case, the policy needs to make fine adjustments to make sure the T is fully in the goal region before heading to the end-zone. 

\subsubsection{Results and Comparison}
As shown in Fig.~6, FODMP achieves the highest success rate on the Push-T task, outperforming DP, ManiCM, and MPD by margins of 19.7\%, 23.6\%, and 9.2\%, respectively. Methods such as DP and ManiCM suffer from temporally unstable action sequences, which lead to accumulated errors over long horizons. In contrast, FODMP maintains temporally consistent motion while enabling fast, one-step inference, resulting in more stable contact dynamics and significantly improved task success.
Beyond final performance, FODMP also demonstrates superior data efficiency on the Push-T task. As shown in Fig.~\ref{fig:data-effiency}(b), FODMP achieves high success rates with substantially fewer demonstrations, while baseline methods require significantly more data to reach comparable performance.

\subsection{RealWorld Ball Catching Experiments}
\begin{figure}
    \centering
    \includegraphics[width=\linewidth]{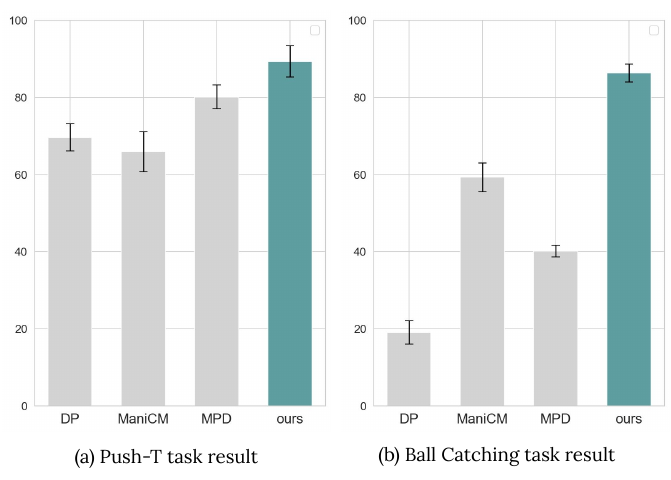}
    \caption{\textbf{RealWorld Experiments Results. }Success rates of different methods on (a) the Push-T task and (b) the Ball Catching task.}
    \label{fig:result}
\end{figure}

The Ball Catching task evaluates the robot’s ability to generate motions for intercepting a dynamically moving object. In this task, a human throws a ball toward the robot with varying trajectories, requiring rapid, closed-loop responses based on real-time observations. In contrast to the Push-T task, which focuses on stable contact manipulation with a relatively static object, Ball Catching emphasizes low-latency inference and temporally coherent motion generation under highly dynamic conditions. Motion discontinuities or delayed responses can directly lead to missed interceptions or unstable interactions. As a result, this task serves as a challenging benchmark for evaluating time-consistent motion generation in real-world dynamic scenarios.
\begin{figure}
    \centering
    \includegraphics[width=\linewidth]{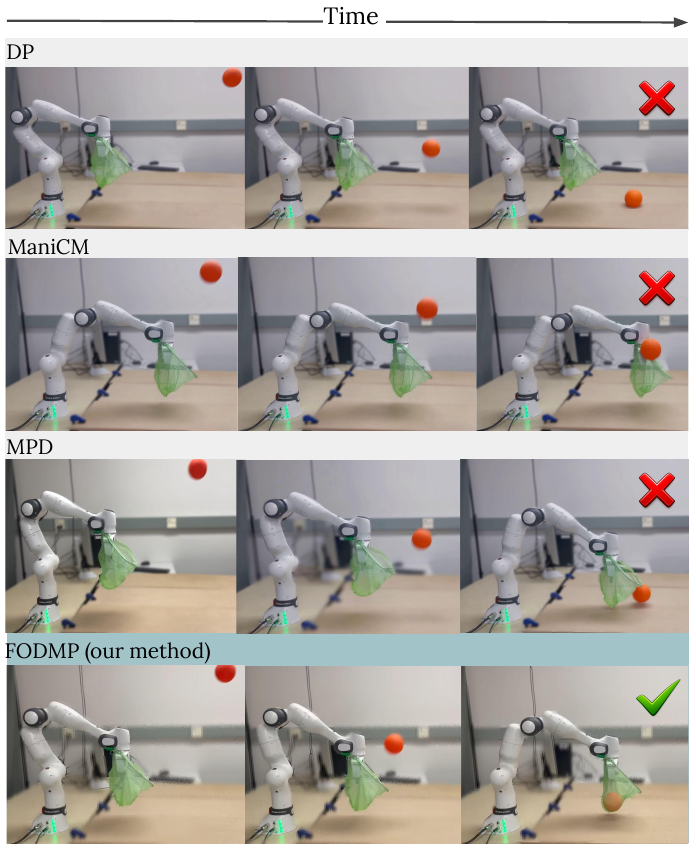}
    \caption{\textbf{Ball Catching Task Performance Comparison. }Snapshots over time illustrate that baseline methods fail to adapt to the fast-moving ball, whereas FODMP produces temporally consistent motions that successfully catch and stabilize the ball.}
    \label{fig:placeholder}
\end{figure}
\subsubsection{Hardware Setup}
As shown in Fig.~\ref{fig:hardware}, we use a Franka Emika Panda robot equipped with a lightweight net as the end-effector. A D455 RealSense depth camera provides real-time depth observations of the incoming ball. In addition, to enhance training stability, we augment the observation space with the estimated ball position and velocity.

\subsubsection{Evaluation Methodology}The Ball Catching task is sensitive to both the thrower and the throwing direction, which can significantly affect task difficulty and evaluation outcomes. To ensure a fair and systematic evaluation, we vary both the human throwers and the throwing directions during experiments. During evaluation, ball throws are categorized into three difficulty levels—easy, medium, and hard—based on the throw direction and trajectory relative to the robot. For each evaluation setting, multiple human participants perform ball throws, with each participant executing 60 throws from different directions, evenly distributed across the three difficulty categories (20 trials per category). This protocol ensures balanced coverage of throw directions across participants and mitigates performance bias arising from individual throwing styles or viewpoints.

\subsubsection{Success Rate Comparison}As shown in Fig.~\ref{fig:result}, FODMP significantly outperforms all baselines on the Ball Catching task, achieving a success rate improvement of 68.2\%, 26.1\%, and 36.2\% over DP, ManiCM, and MPD, respectively. This substantial margin highlights the importance of low-latency inference and time-consistent motion generation in dynamic interception tasks. In contrast, methods that produce temporally inconsistent or delayed actions frequently fail to align the end-effector motion with the ball trajectory, leading to unsuccessful catches.

\subsubsection{Data Efficiency Comparison}
\begin{figure}
    \centering
    \includegraphics[width=\linewidth]{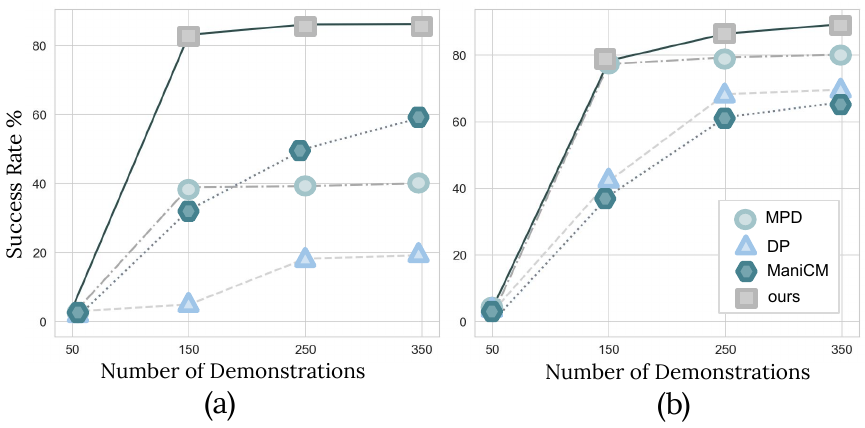}
        \caption{\textbf{Data Effiency Comparison. }(a) Ball Catching task and (b) Push-T task. Success rates of different methods under varying numbers of training demonstrations.}
    \label{fig:data-effiency}
\end{figure}

We collect a total of 350 demonstrations across three difficulty levels (150 easy, 150 medium, and 50 hard). To evaluate data efficiency, all methods are trained multiple times using varying numbers of demonstrations, while maintaining the same proportion of difficulty categories. As shown in Fig.~\ref{fig:data-effiency}(a), FODMP consistently achieves higher success rates across all data regimes, with particularly strong performance in low-data settings. With fewer demonstrations, FODMP exhibits a steeper improvement curve compared to baseline methods, indicating more effective utilization of limited training data.
\subsubsection{Motion Quality}
\begin{figure}[htbp]
    \centering
    \includegraphics[width=\linewidth]{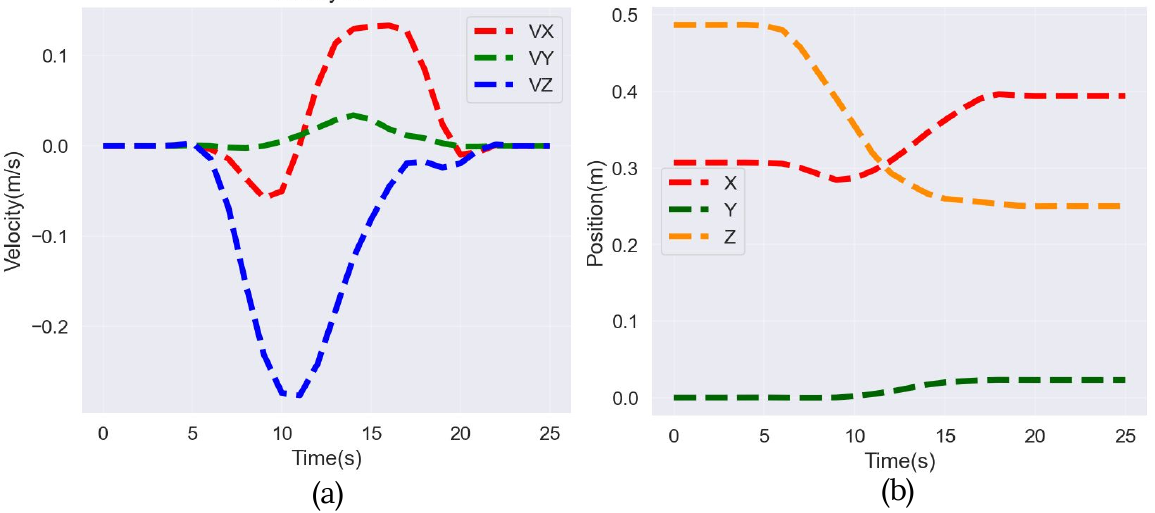}
    \caption{\textbf{Motion analysis. }The temporal evolution of (a) end-effector velocities and (b) positions along Cartesian axes. The smooth velocity transitions and continuous spatial path indicate temporally consistent motion generation}
    \label{fig:trajectory}
\end{figure}

Fig.~\ref{fig:trajectory} illustrates the end-effector motion generated by FODMP, including velocity profiles and positional evolution. The velocity profiles along all axes are smooth and temporally consistent, exhibiting no abrupt changes or oscillations, which indicates stable action generation over time. Correspondingly, the end-effector position evolves continuously, producing a smooth and coherent spatial trajectory. These results demonstrate that FODMP generates high-quality motions that are essential for reliable execution in real-world manipulation tasks.

\section{Conclusions and Limitations}
In this work, we propose FODMP, a novel consistency-distilled movement primitives model that integrates Consistency Models with Probabilistic Dynamic Movement Primitives (ProDMPs~\cite{paraschos2013probabilistic}) for efficient, smooth robotic motion generation. Using Movement Primitive Diffusion (MPD~\cite{scheikl2024movement}) as the teacher model, our approach learns to produce structured motion trajectories while dramatically improving inference speed via consistency distillation. Unlike conventional diffusion policies that require multiple denoising steps, FODMP achieves real-time, single-step inference without sacrificing trajectory quality. In future work, we plan to extend FODMP to complex, high-dimensional tasks and integrate task-specific cost functions into the consistency training framework to further evaluate its applicability in more robotic tasks.

Due to time limits, we did not test our method on large-scale task pre-training using large datasets and large parameterized networks. It's worth noting that our primary contribution --- the design of the action decoder --- is inherently modular and can be seamlessly integrated into large-scale diffusion-based VLA (vision-language-action) models~\cite{dasari2024ingredients, black2024pi_0, shi2025hi, ze20243d}. By employing our method as the action decoder (``action expert'') and pairing it with a powerful encoders, such as the one used in the $\pi_0$~\cite{black2024pi_0} model, our framework holds significant potential for improved performance and scalability across diverse robotic tasks.







\bibliography{fodmp_ref}
\end{document}